\documentclass[letterpaper]{article} 
\usepackage{aaai2026}  
\usepackage{times}  
\usepackage{helvet}  
\usepackage{courier}  
\usepackage[hyphens]{url}  
\usepackage{graphicx} 
\usepackage{natbib}  
\usepackage{caption} 
\usepackage{algorithm}
\usepackage{algorithmic}

\usepackage{newfloat}
\usepackage{listings}
\DeclareCaptionStyle{ruled}{labelfont=normalfont,labelsep=colon,strut=off} 
\floatstyle{ruled}
\newfloat{listing}{tb}{lst}{}
\floatname{listing}{Listing}
\usepackage{amsmath}
\usepackage{amsfonts}
\usepackage{booktabs}
\usepackage{cleveref}
\usepackage{graphicx}
\usepackage{subcaption}
\usepackage{multirow}
\usepackage{bm}

\title{Seeing Right but Saying Wrong: Inter- and Intra-Layer Refinement in MLLMs without Training}
\author{
    Shezheng Song, Shasha Li, Jie Yu
}
\affiliations{
    \textsuperscript{\rm 1}National University of Defense Technology

}

\usepackage{bibentry}

\begin{document}

\maketitle

\begin{abstract}

Multimodal Large Language Models (MLLMs) have demonstrated strong capabilities across a variety of vision-language tasks. However, their internal reasoning often exhibits a critical inconsistency: although deeper layers may attend to the correct visual regions, final predictions are frequently misled by noisy attention from earlier layers. This results in a disconnect between what the model internally understands and what it ultimately expresses, a phenomenon we describe as “seeing it right but saying it wrong.”
To address this issue, we propose DualPD, a dual-perspective decoding refinement strategy that enhances the model’s visual understanding without any additional training. DualPD consists of two components. (1) The layer-wise attention-guided contrastive logits module captures how the model’s belief in the correct answer evolves by comparing output logits between layers that exhibit the largest attention shift. (2) The head-wise information filtering module suppresses low-contribution attention heads that focus on irrelevant regions, thereby improving attention quality within each layer.
Experiments conducted on both the LLaVA and Qwen-VL model families across multiple multimodal benchmarks demonstrate that DualPD consistently improves accuracy without training, confirming its effectiveness and generalizability.
The code will be released upon publication.
\end{abstract}

\section{Introduction}

Multimodal Large Language Models (MLLMs) such as \cite{liu2023llava, Qwen-VL, li2023blip2} have demonstrated impressive capabilities across a wide range of vision-language tasks. However, a fundamental contradiction exists within their internal reasoning process: despite the model acquiring the correct visual knowledge in deeper layers, the final output is often misled by noisy signals from earlier layers. This results in a striking phenomenon: “\textbf{seeing it right but saying it wrong}.” Such misalignment between the model’s internal understanding and its generated response reveals a critical bottleneck that limits MLLMs.

\begin{figure*}
    \centering
    \includegraphics[width=\linewidth]{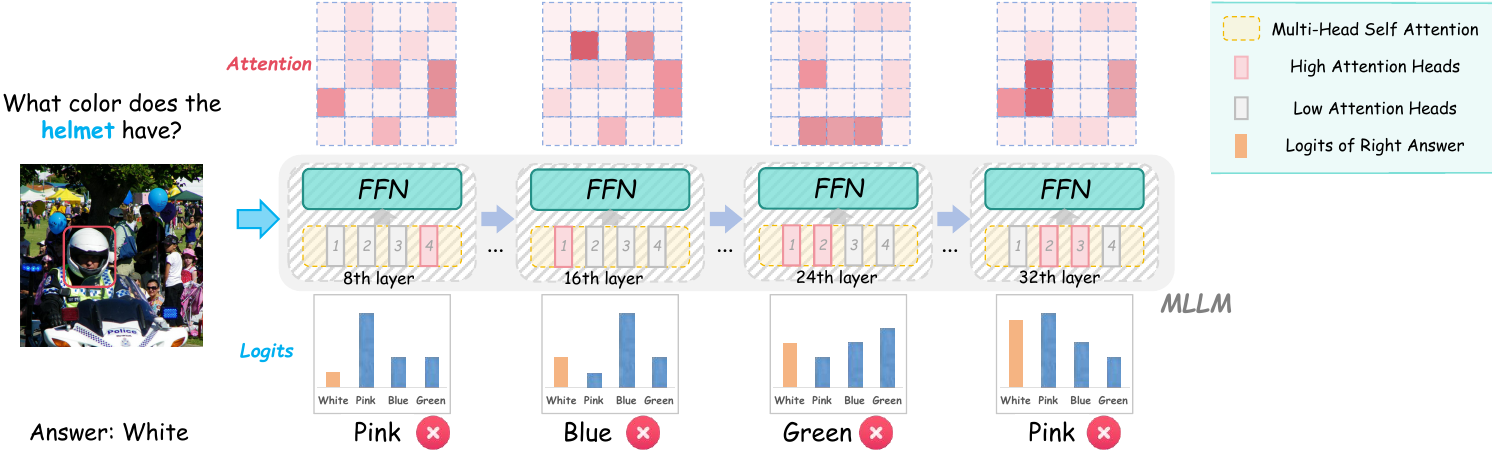}
    \caption{Layer difference for logits and attention.}
    \label{fig:intro_layer}
\end{figure*}

As illustrated in Figure~\ref{fig:intro_layer}, this contradiction is clearly observable. The input image features a person wearing a white helmet, and the correct answer to the question should be “white.” While the attention map from Layer 32 (the final layer) correctly focuses on the helmet, earlier layers such as Layers 8, 16, and 24 show dispersed attention that includes irrelevant regions. This residual influence persists and ultimately causes the model to predict “pink” instead. Interestingly, if we examine the layer-wise output logits, the probability for the correct answer “white” increases steadily with layer depth, suggesting that the model internally “knows” the correct answer. However, the noisy attention from earlier layers inhibits this knowledge from being accurately expressed. 
This internal conflict is not an isolated case; \citet{wang2025exploring} independently confirmed this phenomenon during dataset construction, highlighting it as a fundamental weakness in multimodal reasoning, yet left it unaddressed.
Recent studies have attempted to address similar contradictions by analyzing the evolution of output logits across layers. For example, SLED \cite{zhang2024sled} improves factual consistency by encouraging deeper-layer logits to reflect more reliable knowledge, while DoLA \cite{chuang2023dola} improves generation quality by exploiting differences across intermediate layers. These methods \cite{li2022contrastiveCD, leng2024VCD} effectively mitigate hallucinations from a linguistic perspective, but they largely overlook the \textbf{visual attention trajectory} and its residual interference across layers.  
In order to accurately track the evolution of visual attention, it is also important to consider how attention is distributed within each layer. Attention heads are known to play heterogeneous roles \cite{voita2019analyzing}. Some heads consistently focus on task-relevant regions, while others attend to unrelated content or remain inactive. The effectiveness of each head also varies across layers and tasks. Therefore, treating all heads equally \cite{wu2024controlmllm, lin2024training} may obscure the contribution of more informative heads and weaken the overall attention signal.


To address these challenges, we propose \textbf{DualPD}, a dual-perspective decoding refinement strategy that incorporates both layer-wise and head-wise attention cues. This training-free approach enhances the model's visual understanding by aligning internal attention dynamics:
\textbf{(1) Layer-wise attention-guided contrastive logits}: As the model processes the image through multiple layers, its visual attention gradually shifts from scattered low-level signals to more focused, task-relevant regions. We refer to this process as the \textit{visual attention trajectory}. This evolution is often reflected in the model's output logits across layers, where the confidence for the correct answer becomes more pronounced in deeper layers.
To capture how the model’s visual understanding develops over time, we identify the layer where attention distribution differs most from that of the final layer. We then compare the corresponding output logits between these two layers. This contrast reveals how the model’s internal prediction shifts as visual attention becomes more accurate. In essence, we use the difference in layer-wise logits to approximate what the model is trying to say based on how it looks at the image: tracing how its belief in the correct answer grows as attention becomes more focused.
This strategy enables the logits differences to capture how visual understanding evolves across layers, highlighting key visual cues that guide task-specific predictions.
\textbf{(2) Head-wise information filtering}: To further enhance the accuracy of the visual attention trajectory, we consider the role of individual attention heads within each layer. Some heads consistently focus on informative regions, while others attend to irrelevant areas or remain inactive. We suppress the influence of heads with low attention values, as they are more likely to introduce noise unrelated to the task. This intra-layer filtering helps the model better emphasize useful visual cues, reducing interference from uninformative heads.

Experiments conducted on six multimodal question answering datasets \cite{hudson2019gqa, goyal2017vqav2, singh2019textvqa} demonstrate that our proposed method significantly improves the accuracy. 
In addition, we conduct experiments on multiple models from the LLaVA series \cite{liu2023llava, liu2023improvedllava} to verify our generalizability.
Further attention visualization analyses show that our approach effectively guides MLLM to focus more precisely on task-relevant regions within the image. 
Our main contributions are as follows:
\begin{itemize}
    \item We propose a layer-wise attention-guided contrastive decoding method that traces how the belief in the correct answer evolves during inference. By comparing output logits between layers with the largest shift in attention, this approach captures the trajectory of visual understanding and highlights task-relevant cues.
    \item We introduce a head-wise information filtering mechanism that suppresses the influence of low-attention heads within each layer. This intra-layer adjustment reduces irrelevant signals and enhances the precision of visual reasoning without requiring any additional training.
    \item The proposed method requires no additional training and demonstrates strong generality and effectiveness. It consistently improves the performance of LLaVA and Qwen-VL models across six benchmark datasets.
\end{itemize}

\section{Related Work}

\begin{figure*}
    \centering
    \includegraphics[width=\linewidth]{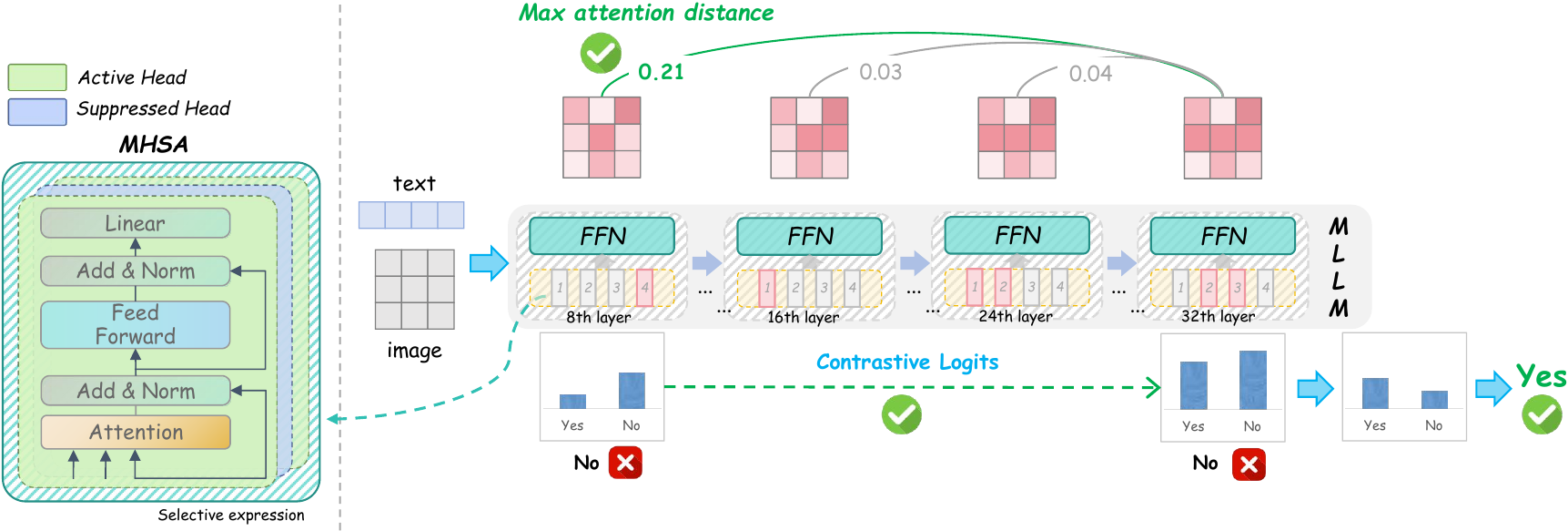}
    \caption{Overview of our framework: inter-layer contrastive attention and intra-layer attention refinement.
    The model first identifies a \textit{basic layer} by selecting the one with the maximum visual attention distance to the \textit{final layer}. It then computes the contrastive logits between the two layers to capture visual information evolution. The resulting logits are used for output prediction.
    The MHSA module (left) performs selective head masking, suppressing low-attention heads to reduce noise. 
    }
    \label{fig:model}
\end{figure*}


Recent decoding-time methods have leveraged contrastive techniques to suppress hallucinations and reduce noise in both language-only and multimodal large models. In the language domain, Contrastive Decoding (CD) \cite{li2022contrastiveCD} mitigates uncertain outputs by contrasting the logits of a large expert model with those of a smaller amateur model, encouraging consistency and discouraging implausible generations. This idea has been extended to vision-language models through approaches such as Visual Contrastive Decoding (VCD) \cite{leng2024VCD}, which contrasts outputs from original and visually perturbed inputs to reduce object hallucinations. Other recent variants, including Language-Contrastive Decoding (LCD) \cite{sennrich2023LCD} and Instruction-Contrastive Decoding (ICD) \cite{wang2024ICD}, further explore contrastive guidance by leveraging alternative prompts or image-conditioned hints \cite{xiao2024seeing}. 
Although these methods are training-free and effective in mitigating noisy or inaccurate predictions, most of them require multiple forward passes through the model. For example, CD \cite{li2022contrastiveCD} involves two passes, and visual cropping \cite{zhang2025mllms} strategies adopted in LLaVA-style models typically require three. This added computational cost significantly increases inference time and limits the practicality of such methods in latency-sensitive scenarios. In contrast, our method refines the decoding process from both the layer-wise and head-wise perspectives using only a forward pass. 

Besides, a growing body of research has explored attention head masking to improve model efficiency, interpretability, and robustness. Recent work \cite{hanheads} investigates the functional roles of attention heads and proposes masking strategies to activate beneficial pathways, though it is designed for text scenario and relies on learned policies. In the multimodal domain, approaches like Learnable Attention Mask (LAM) \cite{barrios2024LAM} apply token-level masking to enhance focus on relevant image regions. Similarly, automatic channel pruning \cite{lee2024automatic} evaluates activation-based redundancy within multi-head attention. Although effective, these methods do not operate directly at the head level, and many require additional training or are limited in generalizability. In contrast, our approach performs head-wise importance filtering entirely during inference by identifying low-attention heads, and suppressing their influence through soft masking. This training-free design offers a lightweight solution for reducing visual noise without modifying model weights or increasing computational cost.

\section{Method}

    We first compute the inter-layer contrastive logits in \cref{sec:inter_layer}, which captures the semantic shift between the basic and target layers. Then, following the intra-layer refinement strategy described in \cref{sec:intra_layer_attention}, we apply head-wise importance filtering by softly suppressing the contributions from low-importance attention heads.
    
\subsection{Problem Formulation}
    Let $(I, T)$ denote the input to a multimodal large language model (MLLM), where $I$ is an image and $T = \{t_1, t_2, \ldots, t_n\}$ is a sequence of text tokens. The model embeds both modalities and processes them through a stack of $L$ transformer layers, each containing $H$ attention heads. The output of the model at each layer is a hidden representation, and the final prediction is generated by applying a linear classifier to the top-layer representation to obtain the logits $\bm{z}^{(L)} \in \mathbb{R}^{|V|}$, where $|V|$ is the vocabulary size.
    
    For clarity, we define the key notations used in our method. Let $\bm{z}^{(l)}$ denote the logits computed from the output of the $l$-th layer. The attention map at layer $l$ is represented as $\bm{A}^{(l)} \in \mathbb{R}^{H \times N \times M}$, where $H$ is the number of attention heads, $N$ is the number of textual tokens, and $M$ is the number of visual tokens. Finally, we denote the predicted answer obtained from the final refined logits as $\hat{y}$.
    The decoding refinement method is composed of two modules: inter-layer contrastive attention and intra-layer attention refine.


\subsection{Inter-layer Contrastive Attention}
\label{sec:inter_layer}

In MLLMs, knowledge and visual understanding are gradually integrated across layers. Shallow layers are responsible for extracting coarse perceptual information, while higher layers are expected to form more abstract and task-relevant representations. However, the standard decoding strategy only uses $\bm{z}^{(L)}$, without explicitly modeling how information evolves across layers.

We propose to capture this evolution by identifying a pair of layers that reflect the model's shift from early perception to deep semantic integration. Specifically, we compute the Hellinger distance between attention maps of adjacent layers to measure attention shift:
    \begin{equation}
        \mathcal{D}_\text{H}(A^{(l)}, A^{(l+1)}) = \sqrt{1 - \sum_{i,j} \sqrt{A^{(l)}_{ij} \cdot A^{(l+1)}_{ij}}}
    \end{equation}
 Then we select the layer $l_b$ with the maximum Hellinger distance as the \textbf{basic layer}, and define the final layer $l_t = L$ as the \textbf{target layer}.
    The inter-layer contrastive logits are computed as:
    \begin{equation}
        \Delta \bm{z} = \bm{z}^{(l_t)} - \bm{z}^{(l_b)}
    \end{equation}

This difference $\Delta \bm{z}$ captures the additional semantic signal introduced between early and late stages of image understanding. It emphasizes the “injected” information and suppresses static or noisy components from early layers, leading to better visual alignment.



\begin{table*}[htb]
  \centering
  \caption{Performance comparison across multimodal benchmarks on LLaVA series.}
    \begin{tabular}{lccccccc}
    \toprule
    \multicolumn{1}{c}{\textbf{Method}} & \textbf{GQA} & \textbf{VQAv2} & \textbf{OKVQA} & \textbf{VizWiz} & \textbf{TextVQA} & \textbf{DocVQA} & \textbf{Average} \\
    \midrule
    \textit{LLaVA-1.5-7B} \cite{liu2023improvedllava} & 66.00  & 76.00  & 51.60  & 57.80  & 57.20  & 24.60  & 55.53  \\
    +DoLA \cite{chuang2023dola} & 68.30  & 73.49  & 51.92  & \textbf{58.30} & 57.00  & 26.40  & 55.90  \\
    +Ours & \textbf{70.60} & \textbf{78.16} & \textbf{53.20} & 57.89  & \textbf{59.30} & \textbf{27.10} & \textbf{57.71} \\
    \midrule
    \textit{LLaVA-1.5-13B} \cite{liu2023improvedllava} & 70.20  & 78.35  & 52.50  & 58.21  & 59.30  & 26.40  & 57.49  \\
    +CD \cite{li2022contrastiveCD}   & 69.10  & 77.56  & 50.21  & 57.87  & 59.60  & 26.00  & 56.72  \\
    +DoLA \cite{chuang2023dola} & 70.20  & 78.96  & 53.52  & \textbf{58.86} & 59.50  & 25.60  & 57.77  \\
    +Ours & \textbf{70.90} & \textbf{79.57} & \textbf{55.04} & 58.12  & \textbf{60.60} & \textbf{27.00} & \textbf{58.54} \\
    \midrule
    \textit{LLaVA-1.6-7B} \cite{liu2023improvedllava} & 69.30  & 79.90  & 52.50  & 59.60  & 72.00  & 65.80  & 66.52  \\
    +CD \cite{li2022contrastiveCD}   & 71.59  & 80.47  & 52.10  & 61.97  & 75.60  & 68.40  & 68.36  \\
    +DoLA \cite{chuang2023dola} & 73.40  & 80.37  & 52.90  & 61.89  & 77.30  & 69.50  & 69.23  \\
    +Ours & \textbf{74.00} & \textbf{81.17} & \textbf{53.00} & \textbf{62.24} & \textbf{77.90} & \textbf{70.50} & \textbf{69.80} \\
    \midrule
    \textit{LLaVA-1.6-13B} \cite{liu2023improvedllava} & 72.00  & 80.24  & 55.00  & 63.11  & 78.20  & 68.10  & 69.44  \\
    +CD \cite{li2022contrastiveCD}   & 72.20  & 80.38  & 54.67  & 63.29  & 77.96  & 68.54  & 69.51  \\
    +DoLA \cite{chuang2023dola} & 71.90  & 80.75  & 54.98  & 63.65  & 78.20  & 69.50  & 69.83  \\
    +Ours & \textbf{73.20} & \textbf{81.29} & \textbf{55.74} & \textbf{64.96} & \textbf{80.10} & \textbf{71.10} & \textbf{71.07} \\
    \bottomrule
    \end{tabular}%
    \label{tab:methods}%
\end{table*}%

\subsection{Intra-layer Attention Refine}
\label{sec:intra_layer_attention}

While the inter-layer mechanism enhances attention flow across layers, it does not address redundancy or noise within a single layer. In practice, attention heads within the same layer often play distinct roles: some focusing on key visual elements, others distracted by background clutter. Thus, we propose a \textbf{intra-layer attention head selection strategy} that suppresses low-importance heads and refines the feature.
We first compute an importance score for each attention head in layer $l$ by measuring the L2 norm of its attention map (prior to softmax normalization):

\begin{equation}
    s_h^{(l)} = \|A^{(l)}_h\|_2
\end{equation}

We then apply a top-$k$ selection strategy:
The top-$k$ heads with highest scores are marked as active.
The remaining $H-k$ heads are suppressed, using a soft decay factor $\gamma$.
Let $\bm{z}^{(l)}_h$ denote the logits produced from head $h$ at layer $l$. We update:

\begin{equation}
    \tilde{\bm{z}}^{(l)}_h = 
        \begin{cases}
        \bm{z}^{(l)}_h & \text{if } h \in \text{Top-}k \\
        \gamma \cdot \bm{z}^{(l)}_h & \text{otherwise}
        \end{cases}
\end{equation}

The refined logits of the layer are then aggregated:
\begin{equation}
    \tilde{\bm{z}}^{(l)} = \sum_{h=1}^{H} \tilde{\bm{z}}^{(l)}_h
\end{equation}

\subsection{Final Decoding}

The refined logits $\bm{z}_{\text{final}}$ are obtained by aggregating the adjusted contributions from all heads. Finally, the predicted answer is generated by selecting the token with the highest value in the refined logits:

\[
\hat{y} = \arg\max(\tilde{\bm{z}}^{(L)})
\]

This two-stage process ensures that the model not only leverages deeper semantic understanding accumulated across layers, but also focuses on the most informative attention pathways within each layer, leading to improved visual alignment and multimodal reasoning performance.

\begin{table*}[htb]
  \centering
  \caption{Performance comparison across multimodal benchmarks on Qwen-VL series.}
    \begin{tabular}{lccccccc}
    \toprule
    \multicolumn{1}{c}{\textbf{Method}} & \textbf{GQA} & \textbf{VQAv2} & \textbf{OKVQA} & \textbf{VizWiz} & \textbf{TextVQA} & \textbf{DocVQA} & \textbf{Average} \\
    \midrule
    \textit{Qwen-2-VL-7B \cite{wang2024qwen2vl}} & 66.38  & 78.09  & 48.20  & 72.73  & 83.34  & 85.56  & 72.38  \\
    +DoLA \cite{chuang2023dola} & 66.14  & \textbf{78.57} & 49.36  & 73.05  & 83.99  & 84.23  & 72.56  \\
    +Ours & \textbf{67.54} & 77.82  & \textbf{52.74} & \textbf{74.72} & \textbf{84.84} & \textbf{85.78} & \textbf{73.91} \\
    \midrule
    \textit{Qwen-2.5-VL-7B \cite{bai2025qwen2.5vl}} & 70.94  & 79.30  & 59.42  & 74.11  & 83.72  & 84.94  & 75.41  \\
    +DoLA \cite{chuang2023dola} & 71.35  & 78.85  & 60.20  & 74.38  & 84.19  & 85.03  & 75.67  \\
    +Ours & \textbf{71.86} & \textbf{80.78} & \textbf{60.45} & \textbf{75.80} & \textbf{85.34} & \textbf{86.80} & \textbf{76.84} \\
    \bottomrule
    \end{tabular}%
  \label{tab:method_qwen}%
\end{table*}%

\begin{table*}[htb]
  \centering
  \caption{Ablation on intra-layer attention refinement (AR, \cref{sec:intra_layer_attention}) and inter-layer contrastive attention (CA, \cref{sec:inter_layer}).}
    \begin{tabular}{clccccccc}
    \toprule
          &       & \textbf{GQA}   & \textbf{VQAv2} & \textbf{OKVQA} & \textbf{VizWiz} & \textbf{TextVQA} & \textbf{DocVQA} & \textbf{Average} \\
    \midrule
    \multirow{3}[2]{*}{\textit{LLaVA-1.5-7B}} & \textbf{Ours}  & \textbf{70.60} & \textbf{78.16} & \textbf{53.20} & \textbf{57.89} & \textbf{59.30} & \textbf{27.10} & \textbf{57.71} \\
          & w/o AR & 69.90  & 77.40  & 52.58  & 57.23  & 58.70  & 25.70  & 56.92  \\
          & w/o AR, CA & 66.00  & 76.00  & 51.60  & 57.80  & 57.20  & 24.60  & 55.53  \\
    \midrule
    \multirow{3}[2]{*}{\textit{LLaVA-1.6-7B}} & \textbf{Ours}  & \textbf{74.00} & \textbf{81.17} & \textbf{53.00} & \textbf{62.24} & \textbf{77.90} & \textbf{70.50} & \textbf{69.80} \\
          & w/o AR & 73.26  & 80.33  & 52.77  & 61.83  & 75.17  & 68.74  & 68.68  \\
          & w/o AR, CA & 69.30  & 79.90  & 52.50  & 59.60  & 72.00  & 65.80  & 66.52  \\
    \bottomrule
    \end{tabular}%
  \label{tab:ablation}%
\end{table*}%

\section{Experiments}

    \subsection{Datasets and Referenced Methods}

    The experiments are conducted on six public multimodal datasets, including GQA \cite{hudson2019gqa}, VQAv2 \cite{goyal2017vqav2}, OKVQA \cite{marino2019okvqa}, VizWiz \cite{gurari2018vizwiz}, TextVQA \cite{singh2019textvqa}, and DocVQA \cite{mathew2021docvqa}. Our evaluations are based on the LLaVA family of models, including LLaVA-1.5-7B, LLaVA-1.5-13B \cite{liu2023llava}, LLaVA-1.6-7B, and LLaVA-1.6-13B \cite{liu2023improvedllava}. We additionally include the Qwen-VL series, covering Qwen2-VL-7B \cite{wang2024qwen2vl} and Qwen2.5-VL-7B \cite{bai2025qwen2.5vl}. The soft decay factor $\gamma$ is set as 0.9.

    We compare our method with the following referenced methods:
    (1) Basic LLaVA decoding \cite{liu2023llava}, which adopts either greedy decoding or sampling strategies.
    (2) Contrastive Decoding (CD) \cite{li2022contrastiveCD}, which uses a small model and a large model for comparison. In the case of LLaMA-based models, CD typically employs LLaMA-7B as the amateur model and LLaMA-13B (or similar) as the expert model. Decoding is guided by the differences between the logits of the two models. Since the CD method requires both a small model and a larger model for comparison, we adopt LLaVA1.5-7B as the small model in the CD setting. As a result, CD can only be applied in comparisons based on larger models such as LLaVA1.5-13B, but not in evaluations where LLaVA1.5-7B itself serves as the primary model.
    (3) DoLA \cite{chuang2023dola}, a training-free inference-time decoding strategy that contrasts output distributions from different layers of the Transformer. It guides the model toward generating more truthful and reliable answers and has shown improved factual accuracy across multiple question answering tasks.


    \subsection{Results and Comparison}


    As shown in Table \ref{tab:methods}, experimental results show that our method outperforms the traditional Contrastive Decoding (CD) approach. Traditional CD methods typically require two large language models to be deployed simultaneously: one small model such as LLaMA-7B and one large model such as LLaMA-33B. This dual-model setup imposes high hardware requirements and limits practicality in real-world applications. In contrast, our method only requires a single model and performs decoding during its standard layer-by-layer inference process, making it more efficient and easier to deploy.
    
    Moreover, our method also outperforms the recently proposed DoLA approach. For example, under the LLaVA-1.5-7B and LLaVA-1.5-13B settings, our method achieves average scores of 57.71 and 58.54, slightly higher than DoLA’s 55.90 and 57.77. More notably, under the stronger LLaVA-1.6-7B and LLaVA-1.6-13B backbones, our method reaches 69.80 and 71.07, outperforming DoLA’s 69.23 and 69.83, respectively. 
    DoLA compares the logits from different layers to capture how knowledge evolves across the model depth. While effective in language-only settings, this approach focuses mainly on internal language model dynamics and overlooks the role of visual information. In multimodal tasks, where images are a key source of information, attention-based analysis better reflects how visual inputs influence model predictions. Our method is specifically designed in multimodal scenario, and the results confirm its adaptability and effectiveness.

    In addition to the LLaVA models, as shown in Table \ref{tab:method_qwen}, we also evaluate our method on the Qwen series, including Qwen2-VL-7B and Qwen2.5-VL-7B. Results show that our method remains effective across different model architectures and consistently outperforms traditional decoding approaches.

    \begin{figure*}[htb]
        \centering
        \begin{subfigure}[t]{0.43\linewidth}
            \includegraphics[width=\linewidth]{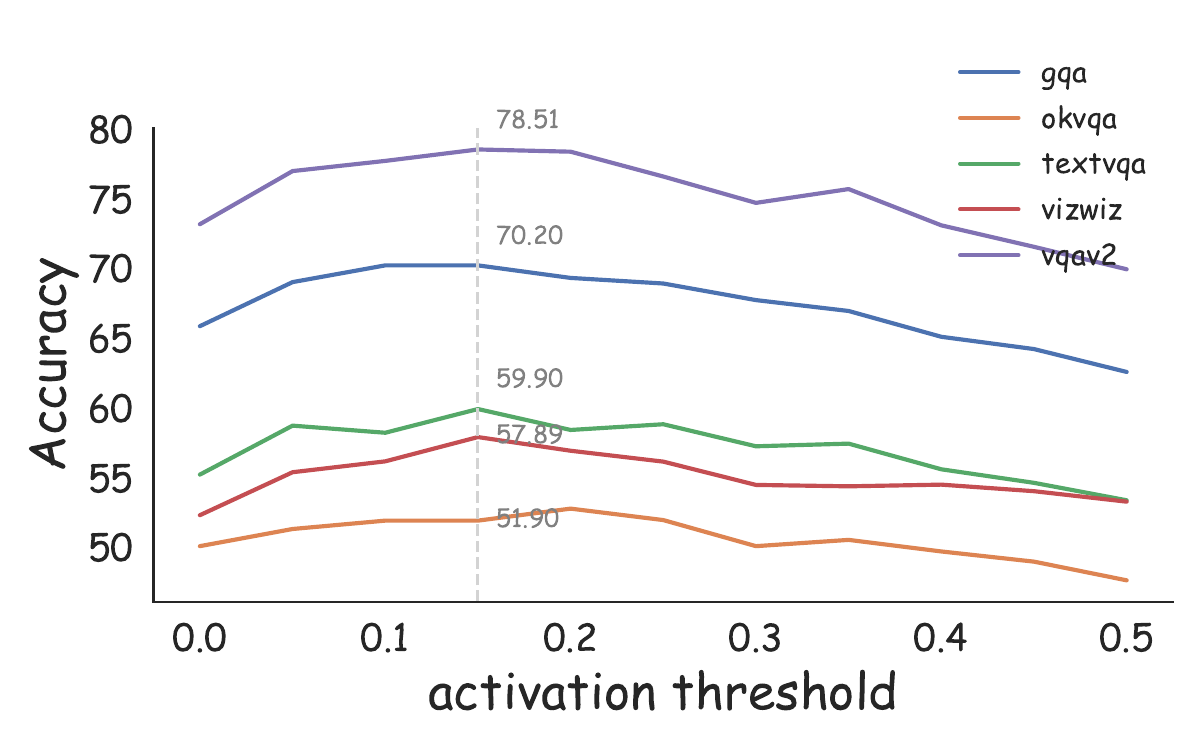}
            \caption{Logit suppression scale = 0.5.}
        \end{subfigure}
        \begin{subfigure}[t]{0.43\linewidth}
            \includegraphics[width=\linewidth]{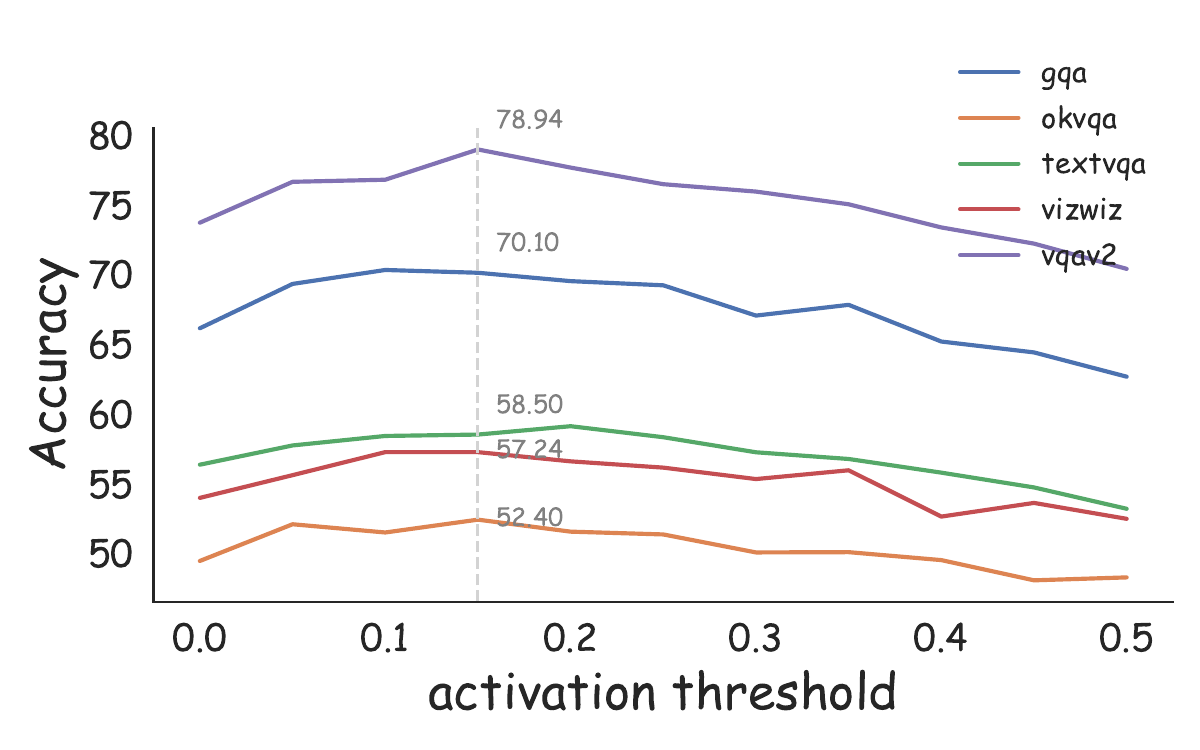}
            \caption{Logit suppression scale = 0.9.}
        \end{subfigure}
    
        \caption{Sensitivity analysis of logit suppression by attention head importance.}
        \label{fig:sensitivityAnalysis}
    \end{figure*}

    \subsection{Ablation Study}

    To evaluate the effectiveness of each component in our proposed method, we perform ablation studies, as summarized in Table~\ref{tab:ablation}. Specifically, we assess the impact of two core modules by selectively removing them from the our method. AR refers to intra-layer attention refinement (\cref{sec:intra_layer_attention}), and CA denotes inter-layer contrastive attention (\cref{sec:inter_layer}).
    (1) \textbf{w/o AR}: This setting removes the intra-layer attention head masking module (\cref{sec:inter_layer}), and applies the inter-layer logits difference $\Delta \bm{z}$ without filtering head-level noise.
    (2) \textbf{w/o AR, CA}: This setting disables both the intra-layer masking and the inter-layer contrastive mechanism (\cref{sec:intra_layer_attention}). It directly uses the final-layer logits $\bm{z}^{(L)}$ for decoding, which corresponds to LLaVA inference.
    
    From the results, we observe the following:
    (1) Removing the intra-layer attention refinement (AR) leads to a noticeable performance drop (from 57.71 to 56.92 average), indicating that selectively suppressing low-contribution heads is effective in reducing noise and improving visual focus.
    (2) When both components are removed (AR, CA), performance further decreases to 55.53, confirming that the inter-layer contrastive attention mechanism also plays a key role in enhancing visual grounding by capturing the semantic transition between layers.
    (3) The full model outperforms both variants across all six datasets, demonstrating that the two modules are complementary. The inter-layer contrast captures high-level attention shifts, while the intra-layer mechanism refines attention locally, leading to more accurate multimodal reasoning.
    These results verify that the proposed component contributes to the overall performance, and their combination enables more precise localization of task-relevant visual information across diverse multimodal tasks.
    
    \subsection{Sensitivity Analysis}
    In multimodal tasks, not all attention heads contribute equally to final predictions. Some heads attend to irrelevant regions, leading to noisy logits. To address this, we suppress the influence of low-attention heads during decoding.
    We perform a sensitivity analysis by varying the activation threshold, which defines the proportion of attention heads with the lowest scores to be suppressed. Specifically, heads below this threshold have their logits scaled (0.5 in (a), 0.9 in (b)). Performance is evaluated across multimodal datasets.
    As shown in Figure \ref{fig:sensitivityAnalysis}, performance consistently peaks when the bottom 15\% attention heads are suppressed (activation threshold = 0.15), across both suppression scales. This suggests that a mild degree of suppression effectively reduces attention noise and improves accuracy. Excessive suppression degrades performance, indicating the importance of preserving moderately relevant heads.

    \subsection{Qualitative Study} 
\begin{table}[htb]
  \centering
  \caption{Case study and comparision between our method and previous representative methods.}
    \resizebox{\linewidth}{!}{ 
    \begin{tabular}{lccc}
    \toprule
    Image & 
    \makebox[1.6cm][c]{\begin{minipage}[c]{0.21\linewidth}\centering{\includegraphics[width=\linewidth]{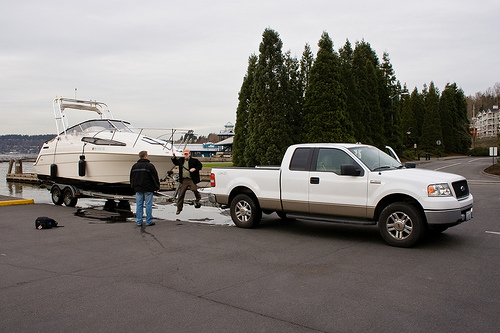}}\end{minipage} }
    & \makebox[1.6cm][c]{\begin{minipage}[c]{0.21\linewidth}\centering{\includegraphics[width=\linewidth]{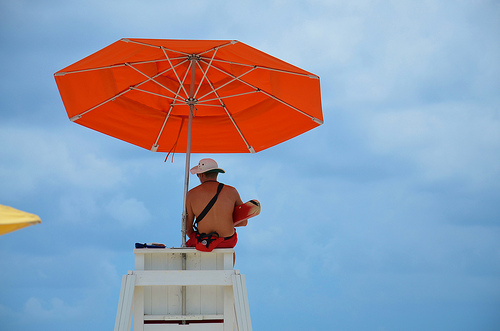}}\end{minipage} }
    & \makebox[1.6cm][c]{\begin{minipage}[c]{0.21\linewidth}\centering{\includegraphics[width=\linewidth]{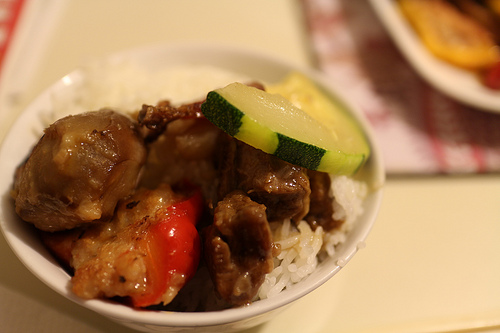}}\end{minipage} } \\
    Text  & \multicolumn{1}{p{1.6cm}}{Is the boat on right?} & \multicolumn{1}{p{1.6cm}}{What color is the large umbrella?} & \multicolumn{1}{p{1.6cm}}{What is right of the beef?}  \\
    \midrule
    LLaVA &  right  $\times$   &  blue $\times$    &   lettuce $\times$  \\
    +DoLA & yes $\times$   & blue $\times$  & broccoli $\times$  \\
    +Ours  & no $\checkmark$   & orange $\checkmark$ & cucumber $\checkmark$ \\
    \bottomrule
    \end{tabular}%
    }
  \label{tab:case}%
\end{table}%

    Figure \ref{tab:case} presents a qualitative comparison among the base LLaVA model, the prior training-free decoding method DoLA, and our proposed method. The base LLaVA fails on all three examples, either providing semantically incorrect answers or misidentifying objects in the visual scene. DoLA slightly improves performance but still produces incorrect results, suggesting its limited capability in precisely locating task-relevant visual cues.
    In contrast, our method successfully answers all questions. In the first example, both LLaVA and DoLA incorrectly predict that the boat is on the right, likely due to misinterpreting spatial relations. Our method correctly determines that the boat is not on the right, showing improved spatial grounding. In the second example, only our method identifies the color of the umbrella as orange, indicating better alignment with visual information. In the third case, our method is the only one to correctly locate the cucumber to the right of the beef, demonstrating more accurate object-level attention.
    These examples clearly demonstrate that our method achieves more accurate visual grounding and spatial reasoning, surpassing both the base model and prior training-free decoding strategies.

    \begin{table*}[htb]
  \centering
  \caption{Comparison with logit-based basic layer selection from DoLA and our attention-based strategy on LLaVA-1.5-7B. As defined in DoLA \cite{chuang2023dola}, we consider three basic layer selection strategies: Low, which limits the candidate layers to 0–15; High, which uses layers 16–31; and All, which allows selection from all layers.}
    \begin{tabular}{lccccccc}
    \toprule
          & \textbf{GQA}   & \textbf{VQAv2} & \textbf{OKVQA} & \textbf{VizWiz} & \textbf{TextVQA} & \textbf{DocVQA} & \textbf{Average} \\
    \midrule
    DoLA Low (0-15) & 68.30  & 73.49  & 51.92  & \textbf{58.30}  & 57.00  & 26.40  & 55.90  \\
    DoLA High (16-31) & 64.30  & 70.11  & 52.80  & 57.40  & 56.10  & 26.00  & 54.45  \\
    DoLA All (0-31)  & 65.40  & 72.97  & 51.34  & 57.77  & 52.50  & 26.10  & 54.35  \\
    \textbf{Ours}  & \textbf{70.60} & \textbf{78.16}  & \textbf{53.20} & 57.89 & \textbf{59.30} & \textbf{27.10} & \textbf{57.71} \\
    \bottomrule
    \end{tabular}%
  \label{tab:method_comparison}
\end{table*}

    \begin{figure*}[htb]
    \centering
    \begin{subfigure}[t]{0.3\linewidth}
        \includegraphics[width=\linewidth]{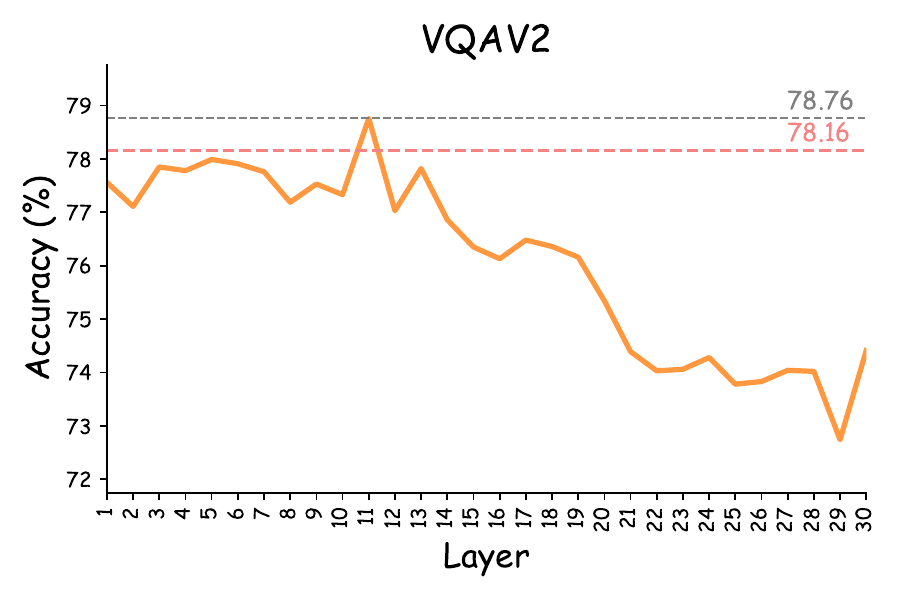}
    \end{subfigure}
    \begin{subfigure}[t]{0.3\linewidth}
        \includegraphics[width=\linewidth]{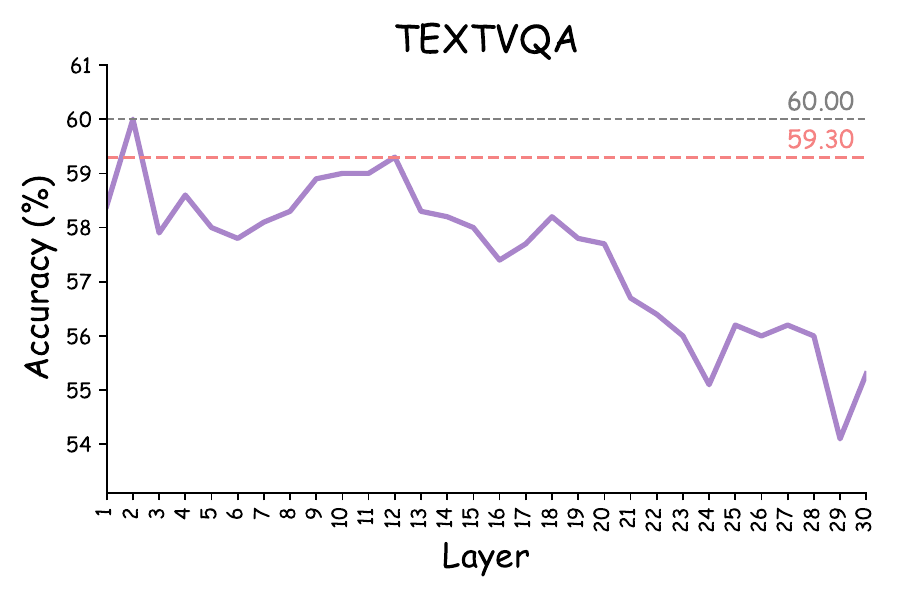}
    \end{subfigure}
    \begin{subfigure}[t]{0.3\linewidth}
        \includegraphics[width=\linewidth]{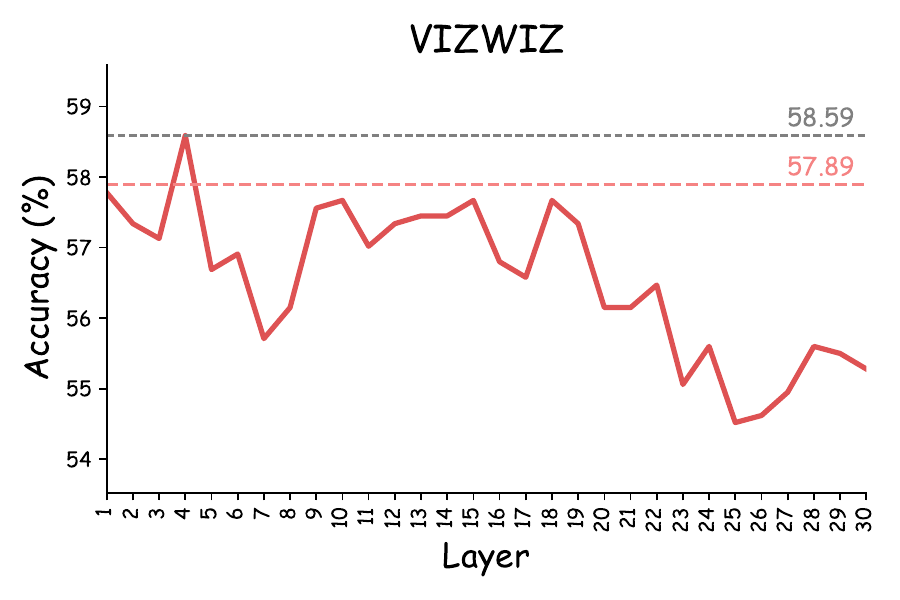}
    \end{subfigure}
    \vspace{0.05cm}
    \\
    \begin{subfigure}[t]{0.3\linewidth}
        \includegraphics[width=\linewidth]{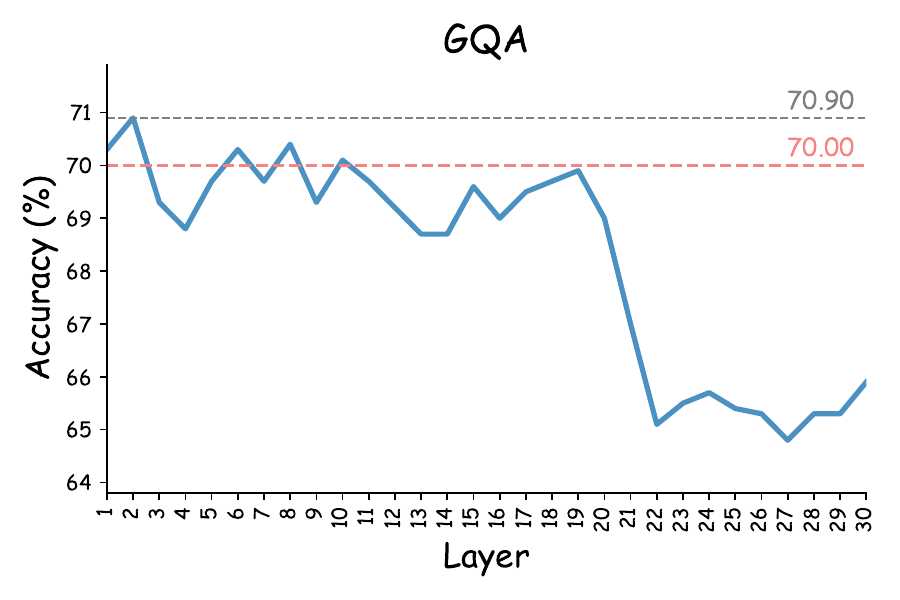}
    \end{subfigure}
    \begin{subfigure}[t]{0.3\linewidth}
        \includegraphics[width=\linewidth]{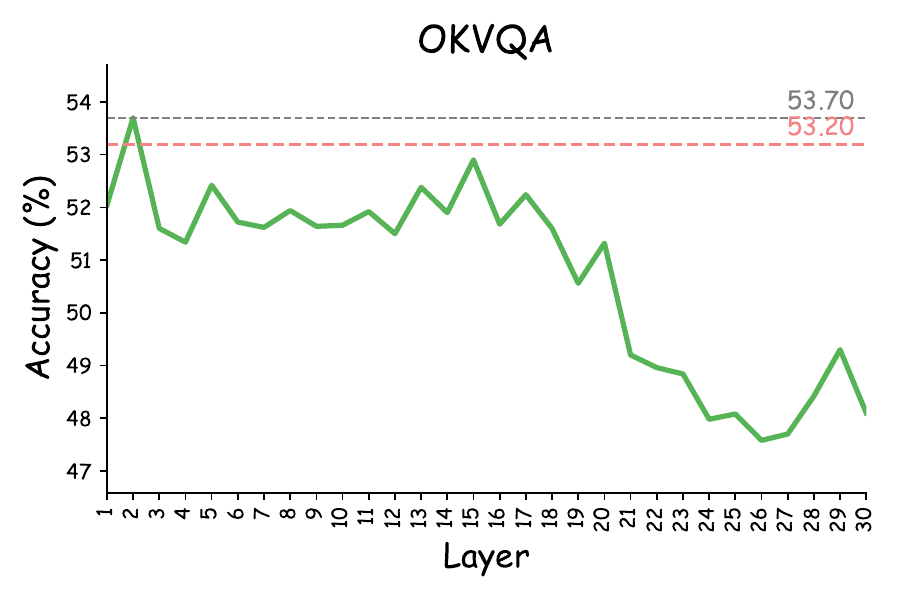}
    \end{subfigure}
    \begin{subfigure}[t]{0.3\linewidth}
        \includegraphics[width=\linewidth]{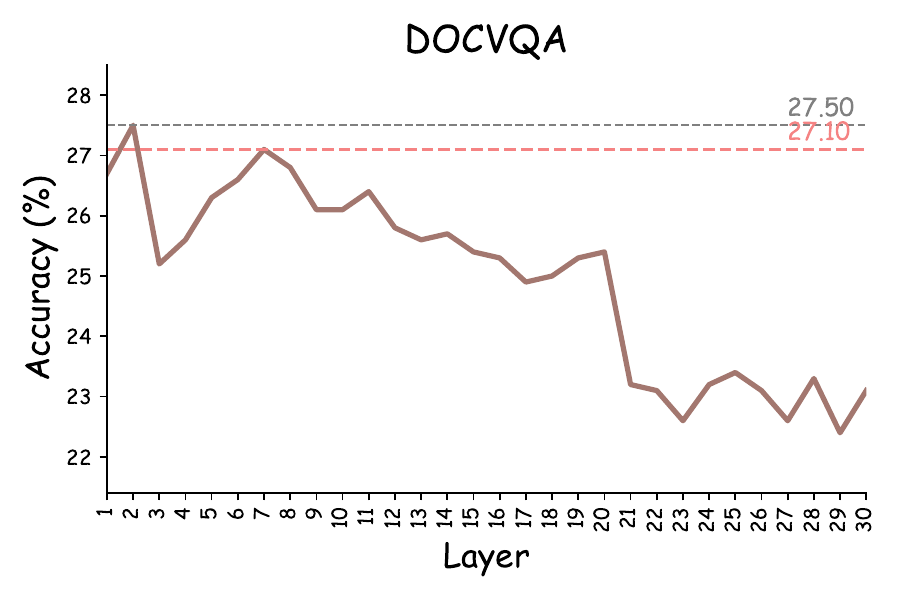}
    \end{subfigure}
    \caption{Static basic layer selection across datasets. The gray dashed line indicates the best accuracy achieved by static selection of a fixed basic layer. The red dashed line represents the accuracy of our method with dynamically selected basic layers based on inter-layer attention difference.}
    \label{fig:static_basic_layer}
\end{figure*}
    \subsection{Basic Layer Seletion}
    \subsubsection{Dynamic Selection}
    As shown in Table \ref{tab:method_comparison}, we compare our method with several layer selection strategies proposed in DoLA. 
    DoLA contrasts logits between basic layer and final layer to enhance factual consistency. 
    While DoLA demonstrated strong performance in language-only settings, we extend and evaluate its basic layer selection mechanisms in the multimodal domain.
    Specifically, we reproduce the three strategies defined in DoLA: Low, which restricts the basic layer to the range of layers 0–15; High, which selects from layers 16–31; and All, which allows selection from 0-31. In these methods, the basic layer is chosen by identifying the one that produces the largest logit difference relative to the final layer. In contrast, our method selects the basic layer by maximizing the difference in visual attention distribution, making it more sensitive to visual signal shifts and better suited for multimodal tasks.
    As the results demonstrate, our approach consistently outperforms all DoLA variants and yields the best average score of 57.71, clearly validating the effectiveness of attention-guided layer selection in multimodal reasoning.

    \subsubsection{Static Selection}
    We further design a static variant of our method to investigate the impact of basic layer selection strategies on performance. In this variant, the basic layer is fixed, allowing us to analyze how performance varies when different layers are used as the basic layer. As shown in Figure \ref{fig:static_basic_layer}, we gradually shift the basic layer from layer 1 to layer 30. The overall trend shows a decline in performance as the layer index increases, indicating that shallower layers are more suitable as the basic layer. This suggests that as the layer depth increases, more visual information is progressively integrated. Using deeper layers as the basic layer may discard these already-fused semantic signals, thereby degrading the final output.

    Although some static layer choices yield slightly higher accuracy than our method in certain datasets, their performance is highly sensitive to the dataset and the specific choice of layer. This makes them less robust and requires careful tuning per task. In contrast, our method dynamically determines the basic layer based on inter-layer visual attention differences, allowing it to generalize better across datasets without manual intervention. As shown in the figure, our approach consistently outperforms most of the static settings, demonstrating its adaptability and effectiveness.
    Notably, the static configuration can be seen as a special case of our approach with a fixed basic layer selection.


\section{Conclusion}
This work identifies a critical inconsistency in the reasoning process of Multimodal Large Language Models (MLLMs), where correct visual understanding in deeper layers is often undermined by noisy attention from earlier stages. To address this issue, we propose DualPD, a dual-perspective decoding refinement strategy that improves visual grounding without any additional training. By leveraging attention-guided contrastive logits across layers and filtering out low-contribution attention heads within each layer, DualPD effectively aligns the model’s internal attention dynamics with its output predictions. Extensive experiments on both the LLaVA and Qwen-VL model families demonstrate that DualPD consistently enhances accuracy, confirming its strong generalizability and adaptability across different MLLM architectures.

\clearpage
\bibliography{aaai25}
\clearpage
\makeatletter
\@ifundefined{isChecklistMainFile}{
  \newif\ifreproStandalone
  \reproStandalonetrue
}{
  \newif\ifreproStandalone
  \reproStandalonefalse
}
\makeatother

\ifreproStandalone
\documentclass[letterpaper]{article}
\usepackage[submission]{aaai2026}
\setlength{\pdfpagewidth}{8.5in}
\setlength{\pdfpageheight}{11in}
\usepackage{times}
\usepackage{helvet}
\usepackage{courier}
\usepackage{xcolor}
\frenchspacing

\begin{document}
\fi
\setlength{\leftmargini}{20pt}
\makeatletter\def\@listi{\leftmargin\leftmargini \topsep .5em \parsep .5em \itemsep .5em}
\def\@listii{\leftmargin\leftmarginii \labelwidth\leftmarginii \advance\labelwidth-\labelsep \topsep .4em \parsep .4em \itemsep .4em}
\def\@listiii{\leftmargin\leftmarginiii \labelwidth\leftmarginiii \advance\labelwidth-\labelsep \topsep .4em \parsep .4em \itemsep .4em}\makeatother

\setcounter{secnumdepth}{0}
\renewcommand\thesubsection{\arabic{subsection}}
\renewcommand\labelenumi{\thesubsection.\arabic{enumi}}

\newcounter{checksubsection}
\newcounter{checkitem}[checksubsection]

\newcommand{\checksubsection}[1]{%
  \refstepcounter{checksubsection}%
  \paragraph{\arabic{checksubsection}. #1}%
  \setcounter{checkitem}{0}%
}

\newcommand{\checkitem}{%
  \refstepcounter{checkitem}%
  \item[\arabic{checksubsection}.\arabic{checkitem}.]%
}
\newcommand{\question}[2]{\normalcolor\checkitem #1 #2 \color{blue}}
\newcommand{\ifyespoints}[1]{\makebox[0pt][l]{\hspace{-15pt}\normalcolor #1}}

\section*{Reproducibility Checklist}

\vspace{1em}
\hrule
\vspace{1em}

\textbf{Instructions for Authors:}

This document outlines key aspects for assessing reproducibility. Please provide your input by editing this \texttt{.tex} file directly.

For each question (that applies), replace the ``Type your response here'' text with your answer.

\vspace{1em}
\noindent
\textbf{Example:} If a question appears as
\begin{center}
\noindent
\begin{minipage}{.9\linewidth}
\ttfamily\raggedright
\string\question \{Proofs of all novel claims are included\} \{(yes/partial/no)\} \\
Type your response here
\end{minipage}
\end{center}
you would change it to:
\begin{center}
\noindent
\begin{minipage}{.9\linewidth}
\ttfamily\raggedright
\string\question \{Proofs of all novel claims are included\} \{(yes/partial/no)\} \\
yes
\end{minipage}
\end{center}
Please make sure to:
\begin{itemize}\setlength{\itemsep}{.1em}
\item Replace ONLY the ``Type your response here'' text and nothing else.
\item Use one of the options listed for that question (e.g., \textbf{yes}, \textbf{no}, \textbf{partial}, or \textbf{NA}).
\item \textbf{Not} modify any other part of the \texttt{\string\question} command or any other lines in this document.\\
\end{itemize}

You can \texttt{\string\input} this .tex file right before \texttt{\string\end\{document\}} of your main file or compile it as a stand-alone document. Check the instructions on your conference's website to see if you will be asked to provide this checklist with your paper or separately.

\vspace{1em}
\hrule
\vspace{1em}


\checksubsection{General Paper Structure}
\begin{itemize}

\question{Includes a conceptual outline and/or pseudocode description of AI methods introduced}{(yes/partial/no/NA)}
yes

\question{Clearly delineates statements that are opinions, hypothesis, and speculation from objective facts and results}{(yes/no)}
yes

\question{Provides well-marked pedagogical references for less-familiar readers to gain background necessary to replicate the paper}{(yes/no)}
yes

\end{itemize}
\checksubsection{Theoretical Contributions}
\begin{itemize}

\question{Does this paper make theoretical contributions?}{(yes/no)}
yes

	\ifyespoints{\vspace{1.2em}If yes, please address the following points:}
        \begin{itemize}
	
	\question{All assumptions and restrictions are stated clearly and formally}{(yes/partial/no)}
	yes

	\question{All novel claims are stated formally (e.g., in theorem statements)}{(yes/partial/no)}
	yes

	\question{Proofs of all novel claims are included}{(yes/partial/no)}
	yes

	\question{Proof sketches or intuitions are given for complex and/or novel results}{(yes/partial/no)}
	yes

	\question{Appropriate citations to theoretical tools used are given}{(yes/partial/no)}
	yes

	\question{All theoretical claims are demonstrated empirically to hold}{(yes/partial/no/NA)}
	yes

	\question{All experimental code used to eliminate or disprove claims is included}{(yes/no/NA)}
	yes
	
	\end{itemize}
\end{itemize}

\checksubsection{Dataset Usage}
\begin{itemize}

\question{Does this paper rely on one or more datasets?}{(yes/no)}
yes

\ifyespoints{If yes, please address the following points:}
\begin{itemize}

	\question{A motivation is given for why the experiments are conducted on the selected datasets}{(yes/partial/no/NA)}
	yes

	\question{All novel datasets introduced in this paper are included in a data appendix}{(yes/partial/no/NA)}
	NA

	\question{All novel datasets introduced in this paper will be made publicly available upon publication of the paper with a license that allows free usage for research purposes}{(yes/partial/no/NA)}
	NA

	\question{All datasets drawn from the existing literature (potentially including authors' own previously published work) are accompanied by appropriate citations}{(yes/no/NA)}
	yes

	\question{All datasets drawn from the existing literature (potentially including authors' own previously published work) are publicly available}{(yes/partial/no/NA)}
	yes

	\question{All datasets that are not publicly available are described in detail, with explanation why publicly available alternatives are not scientifically satisficing}{(yes/partial/no/NA)}
	NA

\end{itemize}
\end{itemize}

\checksubsection{Computational Experiments}
\begin{itemize}

\question{Does this paper include computational experiments?}{(yes/no)}
yes

\ifyespoints{If yes, please address the following points:}
\begin{itemize}

	\question{This paper states the number and range of values tried per (hyper-) parameter during development of the paper, along with the criterion used for selecting the final parameter setting}{(yes/partial/no/NA)}
	yes

	\question{Any code required for pre-processing data is included in the appendix}{(yes/partial/no)}
	yes

	\question{All source code required for conducting and analyzing the experiments is included in a code appendix}{(yes/partial/no)}
	yes

	\question{All source code required for conducting and analyzing the experiments will be made publicly available upon publication of the paper with a license that allows free usage for research purposes}{(yes/partial/no)}
	yes
        
	\question{All source code implementing new methods have comments detailing the implementation, with references to the paper where each step comes from}{(yes/partial/no)}
	yes

	\question{If an algorithm depends on randomness, then the method used for setting seeds is described in a way sufficient to allow replication of results}{(yes/partial/no/NA)}
	yes

	\question{This paper specifies the computing infrastructure used for running experiments (hardware and software), including GPU/CPU models; amount of memory; operating system; names and versions of relevant software libraries and frameworks}{(yes/partial/no)}
	yes

	\question{This paper formally describes evaluation metrics used and explains the motivation for choosing these metrics}{(yes/partial/no)}
	yes

	\question{This paper states the number of algorithm runs used to compute each reported result}{(yes/no)}
	yes

	\question{Analysis of experiments goes beyond single-dimensional summaries of performance (e.g., average; median) to include measures of variation, confidence, or other distributional information}{(yes/no)}
	no

	\question{The significance of any improvement or decrease in performance is judged using appropriate statistical tests (e.g., Wilcoxon signed-rank)}{(yes/partial/no)}
	no

	\question{This paper lists all final (hyper-)parameters used for each model/algorithm in the paper’s experiments}{(yes/partial/no/NA)}
	yes

\end{itemize}
\end{itemize}
\ifreproStandalone
\end{document}
\fi

\end{document}